\documentclass[letterpaper, 10 pt, conference]{ieeeconf}
\usepackage{xurl}
\usepackage[table]{xcolor}
\usepackage{amsmath}
\usepackage{subfigure}
\usepackage{makecell}
\usepackage{multirow}
\usepackage[utf8]{inputenc}
\usepackage[colorlinks=true,citecolor=green,linkcolor=blue,urlcolor=purple]{hyperref}
\usepackage{rotating}
\usepackage{dsfont}
\usepackage{amsfonts}
\usepackage{booktabs}
\usepackage{cite}
\usepackage{bm}
\usepackage{ulem}
\usepackage{marvosym}

\usepackage{mathrsfs}

\newcommand{\ie}{\textit{i.e.}}
\newcommand{\eg}{\textit{e.g.}}

\newcommand{\etal}{\textit{et al.}}
\newcommand{\tabincell}[2]{\begin{tabular}{@{}#1@{}}#2\end{tabular}}
\IEEEoverridecommandlockouts
\title{\LARGE \bf
	D2NT: A High-Performing Depth-to-Normal Translator
}

\author{Yi Feng, Bohuan Xue, Ming Liu, Qijun Chen, and Rui Fan\textsuperscript{\Letter}
	\vspace{-1em}
	\thanks{This work was supported by the National Key R\&D Program of China under Grant 2020AAA0108100, the National Natural Science Foundation of China under Grant 62233013, the Science and Technology Commission of Shanghai Municipal under Grant 22511104500, the Fundamental Research Funds for the Central Universities under Grants 22120220184 and 22120220214, and the Shanghai Municipal Science and Technology Major Project under Grant 2021SHZDZX0100. \textit{(Yi Feng and Bohuan Xue contributed equally to this work.) ({Corresponding author: Rui Fan.})} }
	\thanks{Yi Feng, Qijun Chen, and Rui Fan are with the Robotics \& Artificial Intelligence Laboratory (RAIL), the College of Electronic \& Information Engineering, the State Key Laboratory of Intelligent Autonomous Systems, and Frontiers Science Center for Intelligent Autonomous Systems, Tongji University, Shanghai 201804, P. R. China. (e-mails: {\tt\small fengyi@ieee.org, qjchen@tongji.edu.cn, rui.fan@ieee.org}) }
	\thanks{Bohuan Xue is with the Department of Computer Science \& Engineering, the Hong Kong University of Science and Technology, Hong Kong SAR, P. R. China. (e-mail: {\tt\small bxueaa@ust.hk)}}
	\thanks{Ming Liu is with the Robotics \& Autonomous Systems Thrust of the Systems Hub, the Hong Kong University of Science and Technology (Guangzhou), Nansha, Guangzhou 511400, P. R. China. (e-mail: {\tt\small eelium@ust.hk})}
}

\begin{document}

	\maketitle
	\thispagestyle{empty}
	\pagestyle{empty}

	\begin{abstract}
		Surface normal holds significant importance in visual environmental perception, serving as a source of rich geometric information. However, the state-of-the-art (SoTA) surface normal estimators (SNEs) generally suffer from an unsatisfactory trade-off between efficiency and accuracy. To resolve this dilemma, this paper first presents a superfast depth-to-normal translator (D2NT), which can directly translate depth images into surface normal maps without calculating 3D coordinates. We then propose a discontinuity-aware gradient (DAG) filter, which adaptively generates gradient convolution kernels to improve depth gradient estimation. Finally, we propose a surface normal refinement module that can easily be integrated into any depth-to-normal SNEs, substantially improving the surface normal estimation accuracy. Our proposed algorithm demonstrates the best accuracy among all other existing real-time SNEs and achieves the SoTA trade-off between efficiency and accuracy.
		
	\end{abstract}
	
	\section*{Source Code, Demo Video, \& Supplement}
	Our source code, demo video, and supplement are publicly available at \url{mias.group/D2NT}.

	\section{Introduction}
	\label{sec.intro}
	
	Surface normal is an informative visual feature that has been widely used in a variety of robot environmental perception tasks, \eg, visual odometry \cite{li2020structure, liu2019lpd}, scene parsing \cite{fan2020sne, wang2021dynamic, wang2020applying, fan2019pothole, wang2021sne}, and depth estimation \cite{qi2018geonet, qi2020geonet++}. Due to the requirement for real-time execution in such tasks, surface normal estimators (SNEs) should be both accurate and computationally efficient\cite{fan2021three}.
	
	
	Early geometry-based SNEs compute surface normals via either plane fitting (solvable with energy minimization techniques) or weighted neighboring normal aggregation. 
	However, these SNEs typically have an imbalance between accuracy and speed (see Fig. \ref{fig.tradeoff}). In 2015, Nakagawa {\etal} \cite{nakagawa2015estimating} proposed an efficient SNE, which computes surface normals via the cross product of two orthogonal tangent vectors (hereafter called CP2TV). However, its performance on spatial discontinuities is unsatisfactory as a result of inaccurate observed tangent vectors. Recently, Fan {\etal} \cite{fan2021three} introduced an efficient and accurate SNE, referred to as three-filters-to-normal (3F2N). Although 3F2N achieves state-of-the-art (SoTA) performance, the aggregation of neighboring surface normals with a mean or median filter is still computationally intensive. 
	
	Therefore, there is a strong necessity to develop an SNE that achieves a balance between rapid computation and high accuracy. In this paper, we present a high-performing depth-to-normal translator (D2NT), which significantly improves the efficiency and accuracy trade-off, and significantly refines the estimation results in and around discontinuities. The contributions of our work are summarized as follows:

	\begin{figure}[!t]
		\centering
		\includegraphics[width=0.48\textwidth]{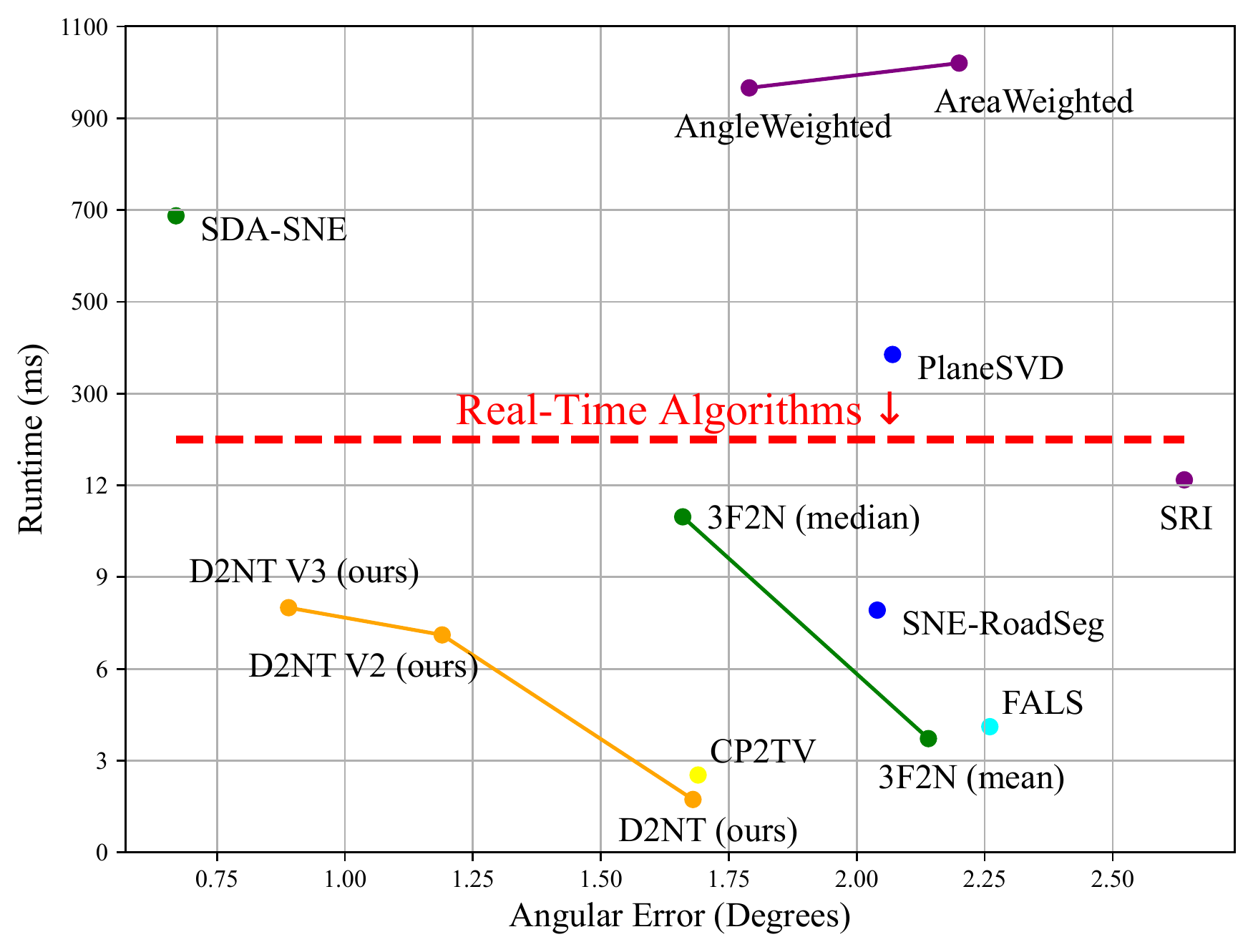}
		\caption{Efficiency versus accuracy trade-off comparison among all SoTA geometry-based SNEs (on the 3F2N Easy dataset). D2NT has the highest computational efficiency, and D2NT V3 achieves the best trade-off between speed and accuracy.} 
	\label{fig.tradeoff}
	\vspace{-1.5em}
\end{figure}

\begin{enumerate}
	\item \uline{\textbf{D2NT}, a cutting-edge \textbf{D}epth-\textbf{to}-\textbf{N}ormal \textbf{T}ranslator}. In comparison to other geometry-based SNEs, D2NT computes surface normals directly from depth maps, demonstrating remarkable computational efficiency. Compared to existing SoTA SNEs, D2NT establishes the most direct relationship between depth and surface normal.
	
	\item \uline{\textbf{D}iscontinuity-\textbf{A}ware \textbf{G}radient (\textbf{DAG}) filter}, a depth gradient filter that selectively identifies discontinuities and eliminates outliers (non-coplanar points in relation to the reference point). Compared to traditional finite difference (FD) operators, our proposed DAG filter provides a significant improvement in terms of depth gradient estimation accuracy.
	
	\item \uline{\textbf{M}arkov random field-based \textbf{N}ormal \textbf{R}efinement (\textbf{MNR}) module}, which dramatically reduces surface normal estimation errors. It can also be integrated with any depth-to-normal SNEs to further enhance the quality of their estimated surface normals. 
	
	%
	
\end{enumerate}


\section{Related work}
\label{sec.related_work}

This section provides an overview of geometry-based surface normal estimators. As shown in Table \ref{table.algos}, the existing SNEs can be divided into three categories: {energy minimization-based, averaging-based, and depth-to-normal}.	

Let $\bm{P}=\{\boldsymbol{p}_1,\boldsymbol{p}_2,...,\boldsymbol{p}_n\}$ be the given 3D point set. For an arbitrary 3D point $\boldsymbol{p}_i\in \bm{P}$, its surface normal is represented as $\boldsymbol{n}_{i}=[n_{ix}, n_{iy}, n_{iz}]^\top$. 
To find the optimal $\boldsymbol{n}_{i}$, $\bm{Q}_i=\{\boldsymbol{q}_{i1}, \boldsymbol{q}_{i2},...,\boldsymbol{q}_{ik}\mid \boldsymbol{q}_{ik}\in \bm{P}\}$, the neighboring points of $\boldsymbol{p}_i$ are typically considered. 

\subsection{Energy Minimization-Based Methods}
This category  of methods computes surface normals by finding a best-fit plane from the augmented neighboring point set $\bm{Q}{_i^+}=\{\bm{Q}_i,\boldsymbol{p}_i\}$ as follows:
\begin{equation}
	\hat{\boldsymbol{n}}_i = \underset {\boldsymbol{n}_i} { \operatorname {arg\,min} } \  
	E(\bm{Q}{_i^+},\boldsymbol{n}_i),
\end{equation}
where $\hat{\boldsymbol{n}}_i$ is obtained by minimizing the energy function $E$.

PlaneSVD \cite{wang2001comparison} fits a local planar surface to $\bm{Q}{_i^+}$ by minimizing the least squares of the distances from the points to the surface using SVD. Similarly, PlanePCA \cite{klasing2009realtime} finds the minimum variance of $\bm{Q}{_i^+}$ with respect to the centroid $\bar{\boldsymbol{p}}_i=\frac{1}{k+1}(\boldsymbol{p}_i+\Sigma_{j=1}^{k}\boldsymbol{q}_{ij})$. VectorSVD \cite{jordan2014quantitative} fits the local planar surface by minimizing the sum of the squared dot products between the surface normal and tangent vectors.

Recently, Ming {\etal} \cite{Ming2022SDA} proposed SDA-SNE, a highly accurate surface normal estimator based on multi-directional dynamic programming and iterative polynomial interpolation. Nevertheless, its demanding computational requirements and iterative nature result in subpar real-time performance. The computation-intensive nature of energy minimization and the calculation of 3D coordinates make these SNEs suffer from slow processing speed and noise.

\subsection{Averaging-Based Methods}
This category of methods estimates surface normals by averaging the normal vectors of the surrounding triangles:
\begin{equation}
	\boldsymbol{n}_i=\frac{1}{k}\sum_{j=1}^k w_j \frac{{\boldsymbol{r}_{i}}_j\times{\boldsymbol{r}_{i}}_{j+1}}{\|{\boldsymbol{r}_{i}}_{j}\times{\boldsymbol{r}_{i}}_{j+1}\|_2},
	\label{eq.averaging_based}
\end{equation}
where $\boldsymbol{r}_{ij}=\boldsymbol{q}_{ij}-\boldsymbol{p}_i$, ${\boldsymbol{r}_{i}}_{k+1}={\boldsymbol{r}_{i}}_{1}$, and $w_j$ is the weight calculated based on either the area (AreaWeighted \cite{klasing2009comparison}) or the angle (AngleWeighted \cite{jin2005comparison}) of the triangles.  Nonetheless, both of these methods necessitate an initial estimation of the normals and can only be utilized as a back-end optimization technique. 

\begin{table}[!t]
	\begin{center}
		\fontsize{7.0}{11.5}\selectfont
		\caption{Taxonomy of the SoTA geometry-based SNEs.}
		\label{table.algos}
		{
			\begin{tabular}{crl}
				\toprule
				Category &Algorithm & Expression  \\
				\hline
				\hline

				\multirow{4}*{\makecell{Energy\\Minimization-\\Based}} 
				&PlaneSVD \cite{wang2001comparison}
				&$\min \Big|\Big|\Big[\bm{Q{_i^+}}, \bm{1}_k\Big]\bm{b_i}\Big|\Big|_2$\vspace{1ex}
				\\

				&PlanePCA \cite{klasing2009realtime}
				&$\min\Big|\Big|\Big[\bm{Q{_i^+}}-\bar{\boldsymbol{p}}\Big]\bm{n_i}\Big|\Big|_2$\vspace{1ex}
				\\

				&VectorSVD \cite{jordan2014quantitative}
				&$\min \Big|\Big| \Big[\bm{Q}_i - \bm{1}_k\bm{p}_i^\top  \Big] \bm{n}_i \Big|\Big|_2$\vspace{1ex}
				\\


				&SDA-SNE \cite{Ming2022SDA}
				&$\min\left\{\bm{\mathcal T}\left(\bm E^{\left(k-1\right)},\bm S \right)\right\}$\vspace{1ex}
				\\
				\hline
				
				\multirow{2}*{\makecell{Averaging-\\Based}} 
				&AreaWeighted \cite{klasing2009comparison}
				&$w_j=\frac{1}{2}\Big|\Big|{\bm{r}_{i}}_{j}\times{\boldsymbol{r}_{i}}_{j+1}\Big|\Big|_2$\vspace{1ex}
				\\

				&AngleWeighted \cite{jin2005comparison}
				&$w_j=\cos^{-1}\Big(\frac{\langle{\boldsymbol{r}_{i}}_{j}, {\boldsymbol{r}_{i}}_{j+1}\rangle}{\|{\boldsymbol{r}_{i}}_{j}\|_2\|{\boldsymbol{r}_{i}}_{j+1}\|_2}\Big)$\vspace{1ex}
				\\
				
				\hline
				
				\multirow{2}*{Depth-to-Normal} 
				&3F2N \cite{fan2021three}
				&\makecell{${n_x=f_x \frac{\partial 1/z}{\partial u},\ \  n_y=f_y \frac{\partial 1/z}{\partial v}}$\vspace{1ex}
					\\
					${\hat{n}_z}=-\Phi\Bigg\{\frac{\Delta {x_i}_j n_x + \Delta {y_i}_j n_y }{\Delta {z_i}_j}\Bigg\}$}\vspace{1ex}
				\\

				&CP2TV \cite{nakagawa2015estimating}
				&$\boldsymbol{n}_i = \bm{t}_u \times \bm{t}_v$
				\\
				
				\bottomrule

			\end{tabular}
		}
	\end{center}
	\vspace{-2em}
\end{table}

\subsection{Depth-to-Normal Methods}
Fan {\etal} \cite{fan2021three} proposed 3F2N, a fast and accurate surface normal estimator, which directly converts the structured range sensor data, such as depth or disparity images, into surface normal maps using two gradient filters and a mean/median filter. This category of methods typically assumes that the range sensor is a pinhole camera model as follows:
\begin{equation}
	z\begin{bmatrix}
		u \\v\\1
	\end{bmatrix}=\bm{K}\bm{p}_i=\begin{bmatrix}
		f_x & 0   & u_\text{o} \\
		0   & f_y & v_\text{o} \\
		0   & 0   & 1
	\end{bmatrix}\begin{bmatrix}
		x \\
		y \\
		z
	\end{bmatrix},
	\label{eq.cam_model}
\end{equation}
where $\bm{K}$ represents the camera intrinsic matrix, $\boldsymbol{p_0}=[u_0,v_0]^\top$ is the principal point in pixels, and $f_x$ and $f_y$ denote the camera's focal lengths in the $x$ and $y$ directions, respectively. This method achieves fast computational speed and high accuracy, but it still involves the calculation of 3D coordinates, which is redundant and computationally demanding.

Nakagawa {\etal} \cite{nakagawa2015estimating} presented CP2TV, an SNE that utilizes the cross-products of tangent vectors of local planar surfaces to directly estimate surface normals from depth maps. However, the accuracy of this method is inadequate in and around discontinuities, as it adopts a finite difference operator to estimate depth gradients. Inaccurate tangent vectors generated in these regions lead to substantial calculation errors in the estimated surface normals.

\begin{figure*}[!htb]
	\centering
	\includegraphics[width=0.99\textwidth]{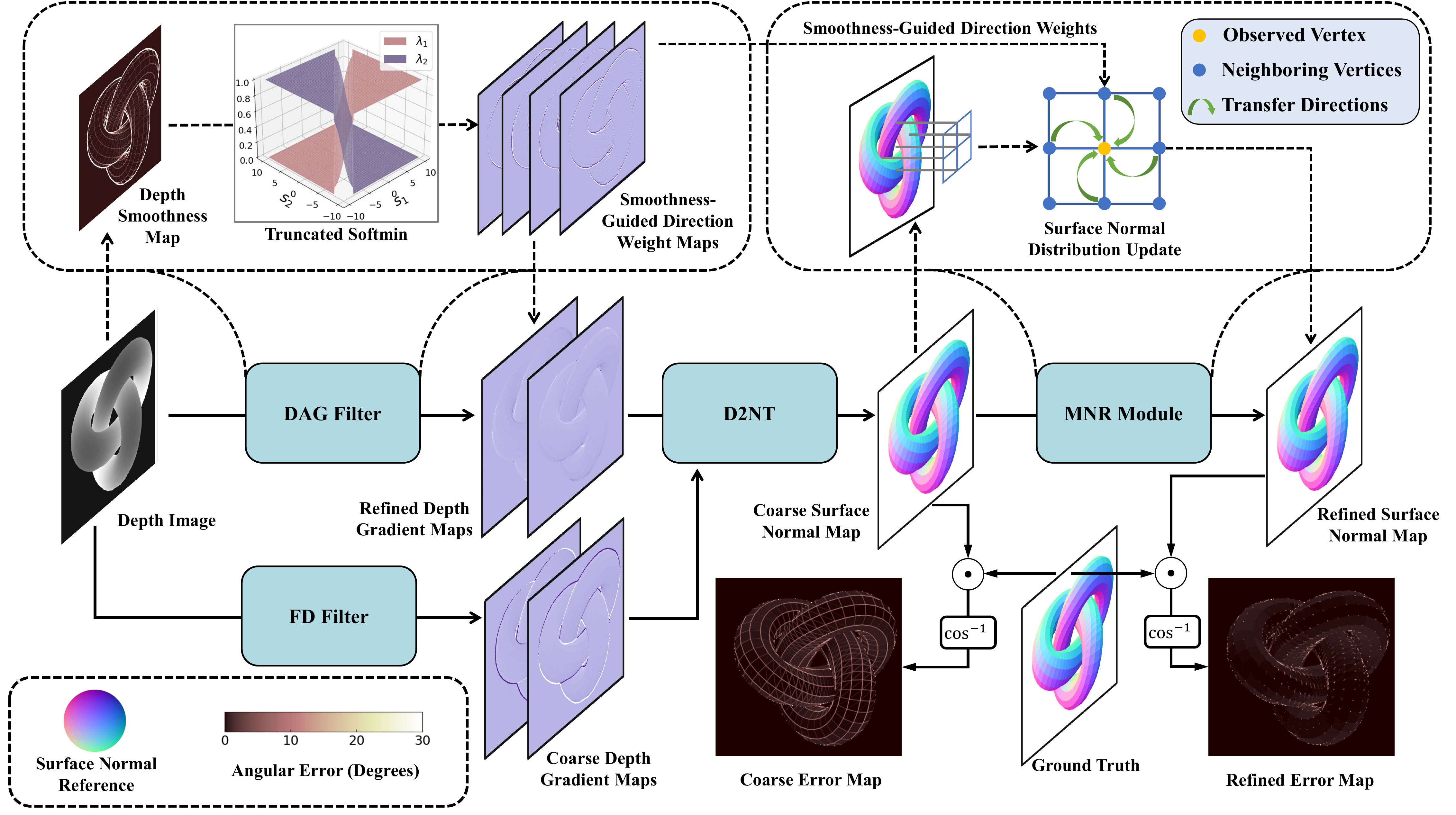}
	\caption{The illustration of our proposed D2NT, DAG filter, and MNR module. D2NT translates depth images into surface normal maps in an end-to-end fashion; DAG filter adaptively generates smoothness-guided direction weights for improved depth gradient estimation in and around discontinuities; MNR module further refines the estimated surface normals based on the smoothness of neighboring pixels.}
	\label{fig.algo_illus}
	\vspace{-1.6em}
\end{figure*}

\section{Methodology}
\label{sec.methodology}

In this section, we first introduce a highly efficient method for estimating surface normals from structured range sensor data in an end-to-end manner. Then, we present a novel approach to improve the accuracy of depth gradient estimation. Additionally, we propose an optimization strategy to refine surface normal estimation in and around discontinuities, which can be well embedded into any existing depth-to-normal SNEs. The pipeline of our algorithm is illustrated in Fig. \ref{fig.algo_illus}.

\subsection{Depth-to-Normal Translation}
\label{subsec.D2NT}

An observed 3D point $\boldsymbol{p}=[x,y,z]^\top$ and its surface normal $\boldsymbol{n}=[n_x, n_y, n_z]^\top$ have the following relation:
\begin{equation}
	\begin{aligned}
		n_{x}x+n_{y}y+n_{z}z+d=0,
	\end{aligned}
	\label{eq.tangent_plane}
\end{equation}
where $d$ is the distance between the origin and the tangent plane. Combining (\ref{eq.cam_model}) with (\ref{eq.tangent_plane}) results in the following expression:
\begin{equation}
	\begin{aligned}
		\frac{z(u-u_0)}{f_x}n_x + \frac{z(v-v_0)}{f_y}n_y + n_zz + d = 0.
	\end{aligned}
	\label{eq.implicit_func}
\end{equation}
(\ref{eq.implicit_func}) contains an implicit function $z(u, v)$. We compute the partial derivatives of $z$ with respect to $u$ and $v$, as follows:
\begin{equation}
	\begin{aligned}
		z_u(\frac{u-u_0}{f_x}n_x + \frac{v-v_0}{f_y}n_y + n_z) + \frac{z}{f_x}n_x = 0, \\
		z_v(\frac{u-u_0}{f_x}n_x + \frac{v-v_0}{f_y}n_y + n_z) + \frac{z}{f_y}n_y = 0,
	\end{aligned}
	\label{eq.patialuv}
\end{equation}
where $z_u = \frac{\partial z}{\partial u}$ and $z_v = \frac{\partial z}{\partial v}$.
$n_x$ and $n_y$ can then be obtained by plugging (\ref{eq.implicit_func}) into (\ref{eq.patialuv}):

\begin{equation}
	\begin{aligned}
		n_x = \frac{f_xd}{z^2}z_u, \ \ \
		n_y = \frac{f_yd}{z^2}z_v.
	\end{aligned}
	\label{eq.nxny}
\end{equation}
$n_z$ can therefore be computed by plugging (\ref{eq.nxny}) into  (\ref{eq.implicit_func}):
\begin{equation}
	\begin{aligned}
		n_z = -\frac{d}{z^2}(z + (u-u_0)z_u + (v-v_0)z_v).
	\end{aligned}
	\label{eq.nz}
\end{equation}
Removing the common factor $\frac{d}{z^2}$ results in a simplified expression for surface normal:
\begin{equation}
	\begin{aligned}
		\boldsymbol{n} = 
		\begin{bmatrix}
			-f_x	&0		&0   \\
			0		&-f_y	&0   \\
			u-u_0	&v-v_0	&z	 \\
		\end{bmatrix}
		\begin{bmatrix}
			z_u   	\\
			z_v   	\\
			1		\\
		\end{bmatrix}.
	\end{aligned}
	\label{eq.mat_n}
\end{equation}
(\ref{eq.mat_n}) describes an end-to-end translation from a given depth image to its surface normal map. Compared with other SoTA SNEs, such as 3F2N \cite{fan2021three} and SNE-RoadSeg \cite{fan2020sne}, D2NT eliminates the need to calculate 3D coordinates by leveraging the explicit relationship between depth and normal, demonstrating remarkable computational efficiency.

\subsection{Discontinuity-Aware Gradient Filtering}
As (\ref{eq.mat_n}) demonstrates, surface normals can be directly calculated from structured range sensor data when the camera parameters are known. The accuracy of the partial derivatives directly affects the accuracy of the surface normal estimation. This subsection introduces an improved depth gradient computation approach.

The existing depth-to-normal methods generally utilize regular image gradient filters, such as FD\footnote{Horizontal FD kernel: $\Delta=[-1,0,1]$.}, to approximate depth gradients. However, these filters tend to yield poor results on discontinuities, such as ridges, ditches, and edges, as outliers (non-coplanar adjacent points) are involved in depth difference computation.

To address this issue, we define a horizontal gradient filter $G_h$ and a vertical gradient filter $G_v$ as follows:
\begin{equation}
	\begin{aligned}
		G_h = \lambda_l\Delta_b + \lambda_r\Delta_f, \ \ \ 
		G_v = \lambda_u\Delta_b^\top + \lambda_d\Delta_f^\top,
	\end{aligned}
	\label{eq.GhGv}
\end{equation}
where $\Delta_b=[-1, 1, 0]$ and $\Delta_f=[0,-1,1]$ are the backward and forward difference operators, respectively. $\lambda$ denotes the weight distribution along four different directions. Hereafter the subscripts $l$, $r$, $u$, and $d$ denote left, right, up, and down directions, respectively. 

To obtain more accurate $G_u$ and $G_v$ in areas with discontinuities, we must distinguish between distinct continuous surfaces and assign appropriate weights to $\Delta_f$ and $\Delta_b$ based on the smoothness of the neighboring pixels' surfaces. The local surface smoothness $s_{\boldsymbol{p}}$ can be reflected by:
\begin{equation}
	\begin{aligned}
		s_{\boldsymbol{p}} = |\nabla^2 z_{\boldsymbol{p}}|,
	\end{aligned}
	\label{eq.smoothness}
\end{equation}
where $\nabla^2$ is a second-order Discrete Laplace Filter (DLF), and $z_{\boldsymbol{p}}$ is the depth at $\boldsymbol{p}$. The weights of the difference operator along four directions can then be assigned as follows:
\begin{equation}
	\begin{aligned}
		\lambda_l, \lambda_r &= \mathscr{M}(s_l, s_r), \ \ \ 
		\lambda_u, \lambda_d &= \mathscr{M}(s_u, s_d),
	\end{aligned}
	\label{eq.assign_lambda}
\end{equation}
where $s_l$, $s_r$, $s_u$, and $s_d$ represent the smoothness of four neighboring points, respectively, and $\mathscr{M}$ is the $\operatorname{softmin}$ function:
\begin{equation}
	\begin{aligned}
		\mathscr{M}(s_i)=\frac{e^{-s_i/\tau}}{\sum_{j=1}^{n} e^{-s_j/\tau}}, \quad  i=1 \ \text{or} \ 2, 
	\end{aligned}
	\label{eq.softmin}
\end{equation}
where $\tau$ is the coefficient that regulates the ``softness'' of the $\operatorname{softmin}$ function. 
Additionally, we observe that significant estimation errors generally occur on the boundaries of surfaces which have small depth gradients (\ie, surfaces that are nearly parallel to the XOY plane). 
This is due to the fact that the adjacent plane typically exhibits a much larger difference in depth gradient magnitude when compared to the reference surface (\textit{i.e.}, the plane where the reference point is located).
As a result, the calculated depth gradient is bound to differ from the depth gradient of the reference plane, even if the weight assigned to the reference plane's depth gradient is high, according to (13).
To tackle this problem, when the smoothness of neighboring points differs greatly from each other, the weights for the depth gradients of adjacent and reference planes should be automatically assigned to 0 and 1, respectively. 
Therefore, we introduce $\operatorname{truncated\ softmin}$ 
\begin{equation}
	\begin{aligned}
		\mathscr{M}_t(s_i)=
		\begin{cases}
			\mathscr{M}(s_i)
			&(|s_2 - s_1|\le 1)\\
			\boldsymbol{\mathds{1}}_{\mathbb{R}^+}{(s_i-1)}
			&(|s_2 - s_1|>1)
		\end{cases}
	\end{aligned}
	\label{eq.tr_softmin}
\end{equation}
to further improve surface normal accuracy,
where $\boldsymbol{\mathds{1}}_{\mathbb{R}^+}(\cdot)$ is the indicator function mapping weight to either 0 or 1 based on the difference in adjacent pixel's surface smoothness. Our proposed DAG filter adaptively generates gradient filters based on surface smoothness, resulting in more accurate estimations of depth gradients, as outliers are effectively filtered out. In summary, the gradient filter of a given point $\boldsymbol{p}_i$ can be represented by the following expression\footnote{Here $\nabla^2(\cdot)$, $|\cdot|$, and $\mathscr{M}_t(\cdot)$ are element-wise operators.}:
\begin{equation}
	\begin{aligned}
		\boldsymbol{G}_i=
		\begin{bmatrix}
			G_h\\
			G_v^\top\\
		\end{bmatrix}_i
		=\mathscr{M}_t(|\nabla^2\boldsymbol{z}_{i}|)
		\begin{bmatrix}
			\Delta_b\\
			\Delta_f\\
		\end{bmatrix},
	\end{aligned}
	\label{eq.filter_mat}
\end{equation}
where $\boldsymbol{z}_{i}=\left[\begin{smallmatrix} z_l & z_r \\ z_u & z_d \end{smallmatrix}\right]$ is the neighborhood depth matrix of $\boldsymbol{p}_i$.

\subsection{MRF-Based Surface Normal Refinement}
Our observation reveals that the surface normals of pixels near/on discontinuities are generally incorrect. This is due to the fact that non-coplanar points are used for local planar surface fitting, causing incorrect depth gradients (discussed in the previous subsection). To resolve this issue, we propose a fast and effective MRF-based optimization (post-processing) method, which significantly improves surface normal accuracy while having minimal impact on the processing speed.




The depth image can be modeled as an undirected graph $\mathcal{G}=(\mathcal{P},\mathcal{E})$, where each node represents a pixel in the depth map, and each edge describes the connection between adjacent pixels. Let $\bm{N}=\{\boldsymbol{n_p}\ \vert \boldsymbol{p}\in \mathcal{P}\}$ be a random variable set, where $\boldsymbol{n_p}$ represents the estimated surface normal of point $\boldsymbol{p}$. $\boldsymbol{n_p}$ is conditionally independent of all other variables in $\bm{N}$:
\begin{equation}
	\boldsymbol{n_p} \perp \!\!\! \perp\boldsymbol{n}_{\mathcal{P}\setminus \bm{Q}{^+_{\boldsymbol{p}}}} \mid \boldsymbol{n}_{\bm{Q}_{\boldsymbol{p}}}.
\end{equation} 
Let $P(\bm{N}=n)$ be the joint probability distribution of $\bm{N}$, representing the probability of a particular field configuration $n$ in surface normal field $\bm{N}$. Specifically, we set the size of the maximum clique to 2 to model our pairwise MRF. According to the Hammersley-Clifford theorem \cite{hammersley1971markov}, $P(\bm{N}=n)$ is represented as follows:
\begin{equation}
	\begin{aligned}
		P(\bm{N}=n)=\frac{1}{Z}\prod_{\boldsymbol{p}\in \mathcal{P}} \phi(\boldsymbol{p})\prod_{(\boldsymbol{p},\boldsymbol{q_i})\in \mathcal{E}}\psi(\boldsymbol{p},\boldsymbol{q_i}),
	\end{aligned}
	\label{eq.PNn}
\end{equation}
where $Z$ is the partition function. (\ref{eq.PNn}) is mathematically equivalent to the energy minimization problem as follows:
\begin{equation}
	\begin{aligned}
		\bm{E}=\sum_{\boldsymbol{p}\in \mathcal{P}} \Phi(\bm{p}) +
		\sum_{(\boldsymbol{p},\boldsymbol{q}_i)\in \mathcal{E}} \Psi(\bm{p},\bm{q}_i),
	\end{aligned}
	\label{eq.Energy}
\end{equation}
where the data term
\begin{equation}
	\begin{aligned}
		\Phi(\boldsymbol{p}) = \vert\vert \boldsymbol{\hat{n}_p}-\boldsymbol{n_p} \vert\vert_2,
	\end{aligned}
	\label{eq.func_phi}
\end{equation} enforces the consistency between the estimated surface normals $\boldsymbol{\hat{n}_p}$ and the observed surface normals $\boldsymbol{n_p}$, and the smoothness term
\begin{equation}
	\begin{aligned}
		\Psi(\boldsymbol{p},\boldsymbol{q}) = s_{\boldsymbol{p}}\sum_{i}\mathscr{M}(\bm{s}_i)\vert\vert\boldsymbol{\hat{n}_p} - \boldsymbol{n}_{\boldsymbol{q}_i}\vert\vert_2,
	\end{aligned}
	\label{eq.func_psi}
\end{equation}
smoothens the surface normal distribution between reference point $\boldsymbol{p}$ and its neighboring point $\boldsymbol{q}_i$, where $\mathscr{M}(\bm{s}_i)$ is the weight of the neighboring point $\boldsymbol{q}_i$ generated by the $\operatorname{softmin}$ function (\ref{eq.softmin}), $\boldsymbol{n}_{\boldsymbol{q}_i}$ is the observed surface normal of $\boldsymbol{q}_i$, and $s_{\boldsymbol{p}}$ is the local surface smoothness that decides the weight between data term and smoothness term.

(\ref{eq.func_phi}) suggests that the difference between the observed and estimated surface normals should be insignificant, while (\ref{eq.func_psi}) implies that adjacent points on the same local planar surface should have consistent normal distributions.

\begin{figure*}[t!]
	\centering
	\includegraphics[width=0.99\textwidth]{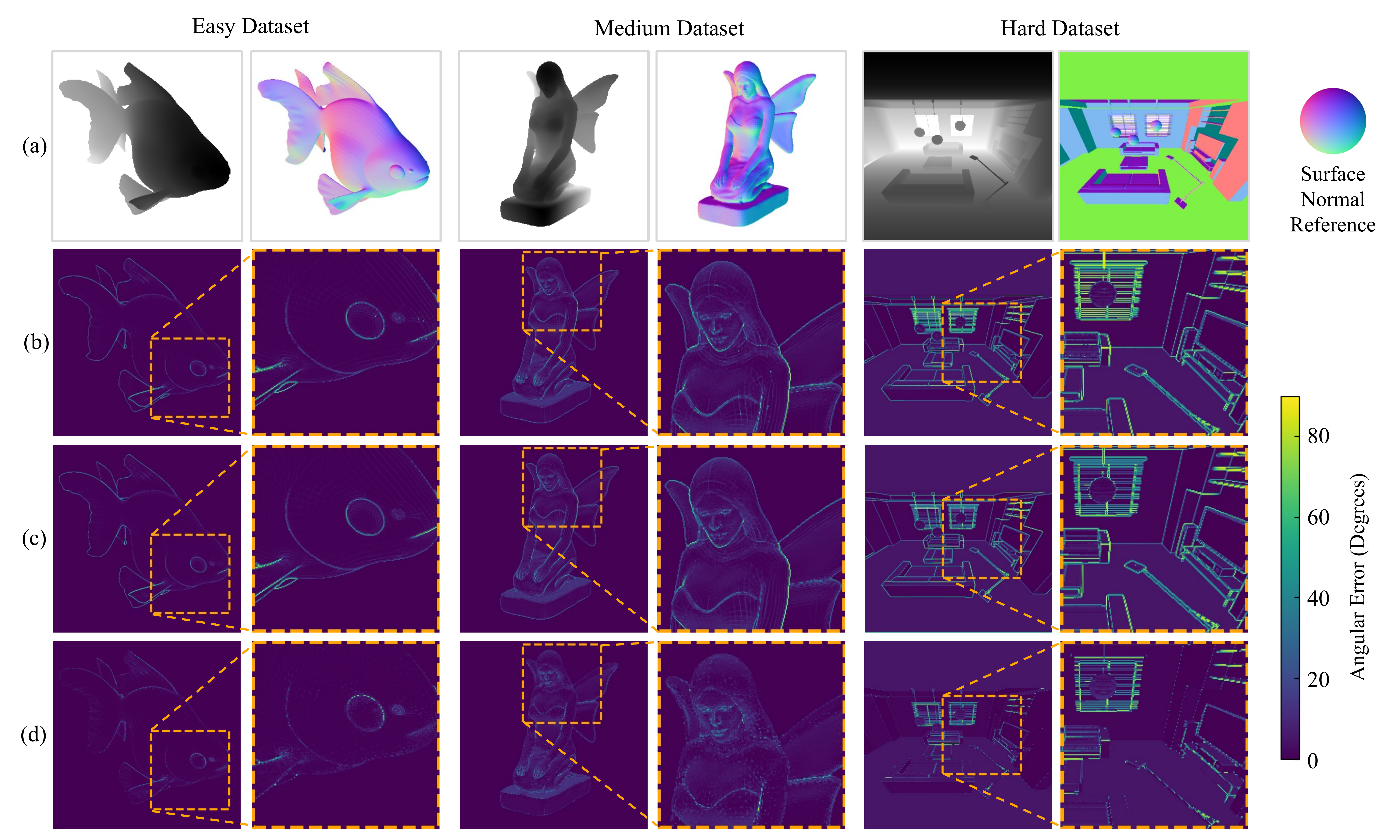}
	\caption{Comparison of our proposed SNE with other SoTA geometry-based SNEs on the 3F2N \cite{fan2021three} dataset: (a) depth maps and ground-truth surface normal maps; (b) error maps obtained using 3F2N (median filter); (c) error maps obtained using CP2TV; (d) error maps obtained using our proposed D2NT V3.}
	\label{fig.comparison}
	\vspace{-1em}
\end{figure*}

\section{Experiments}
\label{sec.experiments}
This section evaluates the performance of our proposed surface normal estimator and compares it with SoTA geometry-based SNEs. To simplify the presentation, we refer to the basic depth-to-normal translator introduced in Sec. \ref{subsec.D2NT} as \uline{\textbf{D2NT}}, the version that includes the DAG filter only as \uline{\textbf{D2NT V2}}, and the version that includes both the DAG filter and the MNR module as \uline{\textbf{D2NT V3}}.


Accurately determining surface normals from real-world range sensor data is infeasible due to the presence of noise. Although public datasets, such as NYUv2 \cite{silberman2012indoor} and DIODE \cite{vasiljevic2019diode}, provide surface normal ``ground truth'', it is often obtained through the interpolation of point sets into local planar surfaces, making the evaluation of SNEs with such ``ground truth'' unreliable. As a result, we conduct experiments on our previously published synthetic dataset \cite{fan2021three}.

\begin{table}[t!]
	\fontsize{7.0}{11.5}\selectfont
	\begin{center}
		\caption{Evaluation of the proposed SNE using four discrete Laplacian filters on the 3F2N datasets.}
		\label{table.kernels}
		{
			\begin{tabular}{c|ccc|ccc}
				\toprule
				\multicolumn{1}{c|}{\multirow{2}*{\tabincell{c}{ Filter\\Config}}}
				&\multicolumn{3}{c|}{D2NT+DAG}
				&\multicolumn{3}{c}{D2NT+MNR}\\
				
				\cline{2-7}
				& Easy 	& Medium 	& Hard 
				& Easy	& Medium  	& Hard \\

				\hline
				\hline
				1D DLF
				&\textbf{1.19}
				&\textbf{4.87}
				&12.84
				&1.19
				&5.08
				&11.87
				\\

				DLF-$\alpha$
				&1.40
				&5.17
				&12.85
				&\textbf{0.79}
				&\textbf{4.80}
				&\textbf{9.86}
				\\

				DLF-$\beta$
				&1.30
				&4.99
				&\textbf{12.46}
				&0.93
				&4.84
				&10.44
				\\

				DLF-$\gamma$
				&1.36
				&5.05
				&13.03
				&1.36
				&5.05
				&13.03
				\\
				
				\bottomrule
			\end{tabular}
		}
	\end{center}
	\vspace{-2.0em}
\end{table}

\begin{table*}[t!]
	\fontsize{7.0}{11.5}\selectfont
	\begin{center}
		\caption{Speed, accuracy, and trade-off comparisons among SoTA geometry-based SNEs on the 3F2N dataset.}
		\label{table.ea_t_pi}
		\begin{tabular}{c|l|r|rrr|rrr}
			\toprule
			\multicolumn{1}{c|}{\multirow{2}*{Real-Time}} 
			& \multicolumn{1}{c|}{\multirow{2}*{Method}} 
			& \multicolumn{1}{c|}{\multirow{2}{*}{$t$ (ms) $\downarrow$}} 
			& \multicolumn{3}{c|}{$e_\text{A}$ (degrees)  $\downarrow$} 
			& \multicolumn{3}{c}{$\pi\ $ (degrees/kHz)  $\downarrow$}\\
			\cline{4-9}
			&  &  &
			\multicolumn{1}{c}{Easy} & 
			\multicolumn{1}{c}{Medium} & 
			\multicolumn{1}{c|}{Hard} & 
			\multicolumn{1}{c}{Easy} & 
			\multicolumn{1}{c}{Medium} & 
			\multicolumn{1}{c}{Hard} \\
			\hline
			\hline
			
			\multirow{6}{*}{\textbf{N}}
			
			& PlaneSVD \cite{klasing2009realtime} 
			& \cellcolor{gray!50}\textbf{393.69}
			& 2.07 
			& 6.07
			& 17.59
			& 813.87
			& \cellcolor{gray!50}\textbf{2389.73}
			& 6923.18\\

			& PlanePCA \cite{jordan2014quantitative}
			& 631.88
			& 2.07
			& 6.07
			& 17.59
			& 1306.29
			& 3835.59
			& 11111.92
			\\
			
			& VectorSVD \cite{klasing2009comparison}     
			& 563.21  
			& 2.13   
			& 6.27   
			& 18.01  
			& 1199.63
			& 3529.11
			& 10142.34
			\\
			& AreaWeighted \cite{klasing2009comparison}  
			& 1092.24  
			& 2.20   
			& 6.27   
			& 17.03  
			& 2407.74
			& 6843.56
			& 18600.68
			\\
			& AngleWeighted \cite{klasing2009comparison} 
			& 1032.88         
			& 1.79          
			& 5.67 
			& 13.26
			& 1850.00       
			& 5855.62       
			& 13693.24
			\\
			
			& SDA-SNE \cite{Ming2022SDA} 
			& 726.18
			& \cellcolor{gray!50}\textbf{0.68}
			& \cellcolor{gray!50}\textbf{4.38}
			& \cellcolor{gray!50}\textbf{8.10}
			& \cellcolor{gray!50}\textbf{493.8}
			& 3180.67
			& \cellcolor{gray!50}\textbf{5882.06}
			\\
			\hline
			
			\multirow{9}{*}{\textbf{Y}}
			
			& SNE-RoadSeg \cite{fan2020sne}  
			& 7.92   
			& 2.04 
			& 6.28 
			& 16.37
			& 16.16
			& 49.74
			& 129.65 
			\\
			
			& 3F2N \cite{fan2021three} 
			& 10.97
			& 1.66 
			& 5.69 
			& 15.31
			& 18.18
			& 62.38
			& 168.03  
			\\
			
			& CP2TV \cite{nakagawa2015estimating} 
			& 2.23
			& 1.69
			& 6.01
			& 13.82
			& 3.75
			& 13.39
			& 30.76
			\\
			
			& D2NT (ours)
			& \cellcolor{gray!80}\textbf{1.82}
			& 1.54
			& 5.64
			& 15.32
			& \cellcolor{gray!80}\textbf{3.05}
			& \cellcolor{gray!80}\textbf{10.25}
			& \cellcolor{gray!80}\textbf{27.84}
			\\
			
			& D2NT V2 (ours)
			&  7.80
			&  1.19
			&  4.87
			&  12.84
			&  8.44
			&  34.67 
			&  91.33\\
			
			& D2NT V3 (ours)
			&  7.99
			&  \cellcolor{gray!80}\textbf{0.89}
			&  \cellcolor{gray!80}\textbf{4.78}
			&  \cellcolor{gray!80}\textbf{9.86}
			&  7.09
			&  38.28
			&  78.91\\
			
			\bottomrule
		\end{tabular}
	\end{center}
	\vspace{-0.5em}
\end{table*}

\begin{table}[t!]
	\fontsize{7.0}{11.5}\selectfont
	\begin{center}
		\caption{Comparison between 3F2N and CP2TV with and without our proposed MNR module embedded.}
		\label{table.with_MRF}
		{
			\begin{tabular}{c|ccc|ccc}
				\toprule
				\multicolumn{1}{c|}{\multirow{2}*{\tabincell{c}{Module\\Config}}}
				
				&\multicolumn{3}{c|}{3F2N}           &\multicolumn{3}{c}{CP2TV}                                                 \\
				
				\cline{2-7}
				

				& Easy 	& Medium 	& Hard 
				& Easy	& Medium  	& Hard \\
				
				\hline
				\hline
				
				w/o MNR
				& 1.66    	& 5.69	& 15.32
				& 1.69    	& 6.02	& 13.82
				\\                          	
				w/ MNR
				& \textbf{0.82}    	& \textbf{4.89}	& \textbf{10.33}  
				&  \textbf{0.91}    & \textbf{4.80}	& \textbf{9.86}  
				\\                          	
				Improvement
				& 50.7\%   	& 14.0\% 	& 32.5\%  
				& 40.8\%   	& 15.0\% 	& 35.6\%   
				\\

				\bottomrule
			\end{tabular}
		}
	\end{center}
\end{table}

\subsection{Implementation Details and Evaluation Metrics}

As discussed in Sec. \ref{sec.methodology}, local surface smoothness is computed through the convolution of the depth map with Laplacian kernels. To find the best convolution kernel, four DLFs are used, including 1D DLF (horizontal kernel: $[1,-2,1]$, and vertical kernel: $[1,-2,1]^\top$), DLF-$\alpha$, DLF-$\beta$, and DLF-$\gamma$\footnote{\scriptsize{DLF-$\alpha$:$\begin{bmatrix}
			0 	& 1		& 0 \\
			1   & -4 	& 1 \\
			0   & 1		& 0
		\end{bmatrix}$, DLF-$\beta$:$\begin{bmatrix}
			1 	& 1		& 1 \\
			1   & -8 	& 1 \\
			1   & 1		& 1
		\end{bmatrix}$, DLF-$\gamma$:$\begin{bmatrix}
			1 	& 2		& 1 \\
			2   & -12	& 2 \\
			1   & 2		& 1
		\end{bmatrix}$}} \cite{wardetzky2020discrete}. 
The execution time of the four DLFs is comparable, as the optimization only occupies a minor portion of the overall process. As demonstrated in Table \ref{table.kernels}, the best results on the 3F2N easy and medium datasets are achieved when using the 1D DLF for D2NT+DAG. Additionally, D2NT+MNR shows the best performance across all three 3F2N datasets when using the DLF-$\alpha$ for computing local surface smoothness. Therefore, we use the 1D DLF for the DAG filter and the DLF-$\beta$ for the MNR module.

Moreover, to meet the real-time requirement, we simplified the implementation of our proposed MNR module. Specifically, when the level of discontinuity in the local surface is assessed to be low according to (\ref{eq.smoothness}), we exclude the smoothness term in (\ref{eq.Energy}). Similarly, if a point is identified to be in and around discontinuities, we omit the data term and instead use the surface normal of the neighboring point with the highest smoothness.

Following \cite{fan2021three}, we use the average angular error $e_\text{A}$ to quantitatively evaluate the performance of SNEs:
\begin{equation}
	\begin{aligned}
		e_{\text{A}}=\frac{1}{N}\sum \limits_{k=1}^{N} \cos^{-1}\left(\frac{\langle{\bm n}_k,\hat{\bm n}_k\rangle}{||{\bm n}_k||_2||\hat{\bm n}_k||_2}\right),
		\label{eq.ea}
	\end{aligned}
\end{equation}
	where $N$ is the number of valid pixels, and $\bm{n}_k$ and $\hat{\bm n}_k$ are the ground truth and estimated surface normals, respectively.
	
	In addition to accuracy evaluation, we adopt the metric 	
	\begin{equation}
		\pi= e_\text{A} t\ (\text{degrees}/\text{kHz})
		\label{eq.pi}
	\end{equation} 
	proposed in \cite{fan2021three} to quantify the trade-off between efficiency and accuracy of a given SNE. A high-performing (fast and accurate) SNE achieves a low $\pi$ score.

	\begin{figure}[!t]
		\centering
		\includegraphics[width=0.48\textwidth]{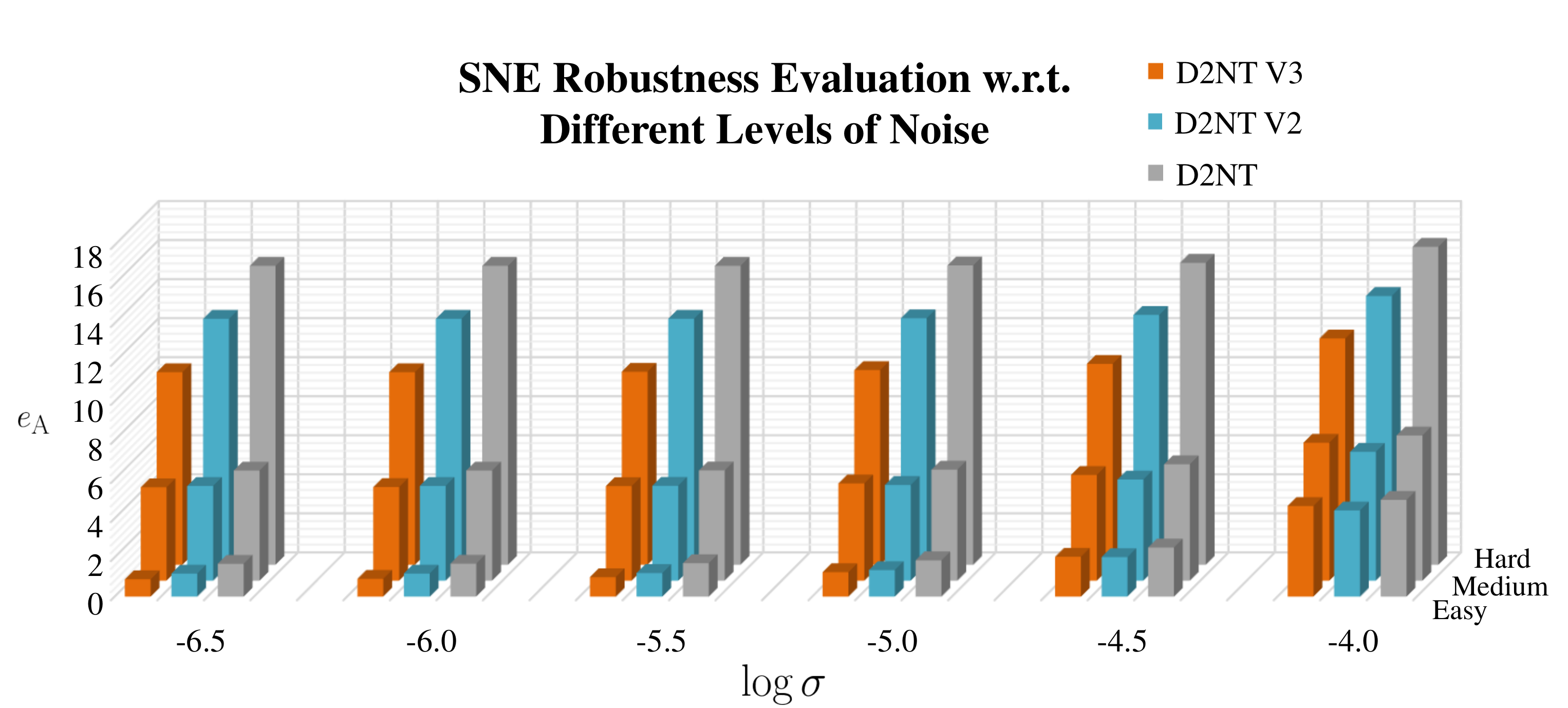}
		\caption{Comparison among the three D2NT versions on the 3F2N datasets with different levels of Gaussian noise added.}
		\label{fig.noise}
	\end{figure}

	\subsection{Performance Comparison}
	As shown in Table \ref{table.ea_t_pi}, our proposed surface normal estimators demonstrate superior performance compared to all other SoTA SNEs. D2NT achieves the highest computational efficiency and the optimum trade-off between speed and accuracy, while D2NT V3 achieves the highest accuracy (the $e_\text{A}$ scores achieved by D2NT are less than $1^\circ$, $5^\circ$, and $9^\circ$ on the 3F2N easy, medium, and hard datasets, respectively). Furthermore, as illustrated in Fig. \ref{fig.comparison}, our D2NT outperforms 3F2N and CP2TV, particularly in and around discontinuities.

	We also conducted supplementary experiments to demonstrate the compatibility of our proposed MNR module with other depth-to-normal SNEs, as shown in Table. \ref{table.with_MRF}. When incorporating the MNR module with 3F2N and CP2TV, the quality of their estimated surface normals is greatly improved, with a drop in 3F2N's $e_\text{A}$ scores by 51\%, 14\%, and 33\% on the 3F2N easy, medium, and hard datasets respectively and a decrease in CP2TV's $e_\text{A}$ scores by 41\%, 15\%, and 36\% on the same datasets. These results suggest that our proposed MNR module can be utilized in conjunction with other depth-to-normal SNEs and serves as an effective back-end optimization technique to enhance surface normal estimation in and around discontinuities.

	As the used synthetic datasets are clean, we further evaluate the robustness of our methods in the presence of random Gaussian noise on the same datasets. As shown in Fig. \ref{fig.noise}, all three D2NT versions are stable with respect to different levels of Gaussian noise, and D2NT V3 is the most robust compared to the other two versions. In addition, it can be observed that our methods exhibit greater stability on the 3F2N medium and hard datasets, as compared to the 3F2N easy dataset. This is likely due to the added discontinuities caused by the Gaussian noise on the 3F2N easy dataset.

	\section{Conclusion}
	This paper presented an end-to-end depth-to-normal translator, a discontinuity-aware gradient filter, and an MRF-based surface normal refinement module. Extensive experimental results demonstrate that 1) our proposed depth-to-normal translator achieves the fastest execution speed and the best balance between computational efficiency and accuracy, and 2) the discontinuity-aware gradient filter and MRF-based surface normal refinement module can further improve its performance in and around discontinuities. Furthermore, our proposed MRF-based surface normal refinement module is also compatible with other depth-to-normal SNEs.

	\clearpage
	\normalem
	\bibliographystyle{IEEEtran}

\begin{thebibliography}{10}
\providecommand{\url}[1]{#1}
\csname url@samestyle\endcsname
\providecommand{\newblock}{\relax}
\providecommand{\bibinfo}[2]{#2}
\providecommand{\BIBentrySTDinterwordspacing}{\spaceskip=0pt\relax}
\providecommand{\BIBentryALTinterwordstretchfactor}{4}
\providecommand{\BIBentryALTinterwordspacing}{\spaceskip=\fontdimen2\font plus
\BIBentryALTinterwordstretchfactor\fontdimen3\font minus
  \fontdimen4\font\relax}
\providecommand{\BIBforeignlanguage}[2]{{%
\expandafter\ifx\csname l@#1\endcsname\relax
\typeout{** WARNING: IEEEtran.bst: No hyphenation pattern has been}%
\typeout{** loaded for the language `#1'. Using the pattern for}%
\typeout{** the default language instead.}%
\else
\language=\csname l@#1\endcsname
\fi
#2}}
\providecommand{\BIBdecl}{\relax}
\BIBdecl

\bibitem{li2020structure}
Y.~Li \emph{et~al.}, ``Structure-{SLAM}: Low-drift monocular {SLAM} in indoor
  environments,'' \emph{IEEE Robotics and Automation Letters}, vol.~5, no.~4,
  pp. 6583--6590, 2020.

\bibitem{liu2019lpd}
Z.~Liu \emph{et~al.}, ``{LPD-Net}: {3D} point cloud learning for large-scale
  place recognition and environment analysis,'' in \emph{Proceedings of the
  IEEE/CVF International Conference on Computer Vision (ICCV)}, 2019, pp.
  2831--2840.

\bibitem{fan2020sne}
R.~{Fan} \emph{et~al.}, ``{SNE-RoadSeg}: Incorporating surface normal
  information into semantic segmentation for accurate freespace detection,'' in
  \emph{European Conference on Computer Vision (ECCV)}.\hskip 1em plus 0.5em
  minus 0.4em\relax Springer, 2020, pp. 340--356.

\bibitem{wang2021dynamic}
H.~Wang \emph{et~al.}, ``Dynamic fusion module evolves drivable area and road
  anomaly detection: A benchmark and algorithms,'' \emph{IEEE Transactions on
  Cybernetics}, vol.~52, no.~10, pp. 10\,750--10\,760, 2022.

\bibitem{wang2020applying}
H.~{Wang} \emph{et~al.}, ``Applying surface normal information in drivable area
  and road anomaly detection for ground mobile robots,'' in \emph{2020 IEEE/RSJ
  International Conference on Intelligent Robots and Systems (IROS)}.\hskip 1em
  plus 0.5em minus 0.4em\relax IEEE, 2020, pp. 2706--2711.

\bibitem{fan2019pothole}
R.~Fan \emph{et~al.}, ``Pothole detection based on disparity transformation and
  road surface modeling,'' \emph{IEEE Transactions on Image Processing},
  vol.~29, pp. 897--908, 2019.

\bibitem{wang2021sne}
H.~Wang \emph{et~al.}, ``{SNE-RoadSeg+}: Rethinking depth-normal translation
  and deep supervision for freespace detection,'' in \emph{2021 IEEE/RSJ
  International Conference on Intelligent Robots and Systems (IROS)}.\hskip 1em
  plus 0.5em minus 0.4em\relax IEEE, 2021, pp. 1140--1145.

\bibitem{qi2018geonet}
X.~{Qi} \emph{et~al.}, ``{GeoNet}: Geometric neural network for joint depth and
  surface normal estimation,'' in \emph{Proceedings of the IEEE Conference on
  Computer Vision and Pattern Recognition (CVPR)}, 2018, pp. 283--291.

\bibitem{qi2020geonet++}
X.~Qi \emph{et~al.}, ``{GeoNet++}: Iterative geometric neural network with
  edge-aware refinement for joint depth and surface normal estimation,''
  \emph{IEEE Transactions on Pattern Analysis and Machine Intelligence},
  vol.~44, no.~2, pp. 969--984, 2020.

\bibitem{fan2021three}
R.~Fan \emph{et~al.}, ``Three-filters-to-normal: An accurate and ultrafast
  surface normal estimator,'' \emph{IEEE Robotics and Automation Letters},
  vol.~6, no.~3, pp. 5405--5412, 2021.

\bibitem{nakagawa2015estimating}
Y.~Nakagawa \emph{et~al.}, ``Estimating surface normals with depth image
  gradients for fast and accurate registration,'' in \emph{2015 International
  Conference on 3D Vision (3DV)}.\hskip 1em plus 0.5em minus 0.4em\relax IEEE,
  2015, pp. 640--647.

\bibitem{wang2001comparison}
C.~Wang \emph{et~al.}, ``Comparison of local plane fitting methods for range
  data,'' in \emph{Proceedings of the IEEE Conference on Computer Vision and
  Pattern Recognition (CVPR)}, vol.~1.\hskip 1em plus 0.5em minus 0.4em\relax
  IEEE, 2001.

\bibitem{klasing2009realtime}
K.~Klasing \emph{et~al.}, ``Realtime segmentation of range data using
  continuous nearest neighbors,'' in \emph{2009 International Conference on
  Robotics and Automation (ICRA)}.\hskip 1em plus 0.5em minus 0.4em\relax IEEE,
  2009, pp. 2431--2436.

\bibitem{jordan2014quantitative}
K.~Jordan \emph{et~al.}, ``A quantitative evaluation of surface normal
  estimation in point clouds,'' in \emph{2014 IEEE/RSJ International Conference
  on Intelligent Robots and Systems (IROS)}.\hskip 1em plus 0.5em minus
  0.4em\relax IEEE, 2014, pp. 4220--4226.

\bibitem{Ming2022SDA}
N.~Ming \emph{et~al.}, ``{SDA}-{SNE}: Spatial discontinuity-aware surface
  normal estimation via multi-directional dynamic programming,'' in \emph{2022
  International Conference on 3D Vision (3DV)}, 2022, pp. 486--494.

\bibitem{klasing2009comparison}
K.~Klasing \emph{et~al.}, ``Comparison of surface normal estimation methods for
  range sensing applications,'' in \emph{2009 International Conference on
  Robotics and Automation (ICRA)}.\hskip 1em plus 0.5em minus 0.4em\relax IEEE,
  2009, pp. 3206--3211.

\bibitem{jin2005comparison}
S.~Jin \emph{et~al.}, ``A comparison of algorithms for vertex normal
  computation,'' \emph{The Visual Computer}, vol.~21, no.~1, pp. 71--82, 2005.

\bibitem{hammersley1971markov}
J.~M. Hammersley and P.~Clifford, ``Markov fields on finite graphs and
  lattices,'' \emph{Unpublished manuscript}, vol.~46, 1971.

\bibitem{silberman2012indoor}
N.~Silberman \emph{et~al.}, ``Indoor segmentation and support inference from
  {RGBD} images,'' in \emph{European Conference on Computer Vision
  (ECCV)}.\hskip 1em plus 0.5em minus 0.4em\relax Springer, 2012, pp. 746--760.

\bibitem{vasiljevic2019diode}
Vasiljevic \emph{et~al.}, ``{DIODE}: A {D}ense {I}ndoor and {O}utdoor {DE}pth
  dataset,'' \emph{CoRR}, 2019.

\bibitem{wardetzky2020discrete}
M.~Wardetzky, ``Discrete laplace operators,'' \emph{An Excursion Through
  Discrete Differential Geometry: AMS Short Course, Discrete Differential
  Geometry, January 8-9, 2018, San Diego, California}, vol.~76, p.~1, 2020.

\end{thebibliography}

\end{document}